%% file: main.tex
\documentclass{article}
\pdfoutput=1

\usepackage[preprint,nonatbib]{neurips_2019}

\usepackage[utf8]{inputenc} 
\usepackage[T1]{fontenc}    
\usepackage{hyperref}       
\usepackage{url}            
\usepackage{booktabs}       
\usepackage{amsfonts}       
\usepackage{nicefrac}       
\usepackage{microtype}      
\usepackage{amsmath}

\usepackage{graphicx}
\usepackage{caption}
\usepackage{color}
\usepackage{wrapfig}

\title{FAN: Focused Attention Networks}
\def\mc{\textsuperscript{1}}
\def\ad{\textsuperscript{2}}
\def\iit{\textsuperscript{3}}
\def\adiit{\textsuperscript{2,3}}
\author{
  Chu Wang \mc \hspace{0.2cm} Babak Samari \mc \hspace{0.2cm} Vladimir G. Kim \ad \hspace{0.2cm} 
  Siddhartha Chaudhuri \adiit \hspace{0.2cm} Kaleem Siddiqi \mc \thanks{Corresponding author.} \\
  \\
  \mc School of Computer Science and Center for Intelligent Machines, McGill University\\
  \ad Adobe Research \\
  \iit Department of Computer Science and Engineering, IIT Bombay \\
  \\
  \texttt{\{chuwang,babak,siddiqi\}@cim.mcgill.ca}, \texttt{\{vokim,sidch\}@adobe.com} \\
}

\begin{document}

\maketitle
\input{tex/abstract.tex}
\input{tex/introduction.tex}

\input{tex/task_limit.tex}

\input{tex/att_sup.tex}
\input{tex/net_architecture.tex}
\input{tex/exp.tex}
\input{tex/conclusion.tex}

\bibliographystyle{plain}
\bibliography{ref}

\newpage
\appendix
\paragraph{\Large Appendix}
\input{tex/supplementary.tex}

\end{document}

%% file: tex/abstract.tex
\begin{abstract}
Attention networks show promise for both vision and language tasks, by emphasizing relationships between constituent elements through weighting functions. Such elements could be regions in an image output by a region proposal network, or words in a sentence, represented by word embedding. Thus far the learning of attention weights has been driven solely by the minimization of task specific loss functions. We introduce a method for learning attention weights to better emphasize informative pair-wise relations between entities. 
The key component is a novel center-mass cross entropy loss, which can be applied in conjunction with the task specific ones. We further introduce a focused attention backbone to learn these attention weights for general tasks. We demonstrate that the focused supervision leads to improved attention distribution across meaningful entities, and that it enhances the representation by aggregating features from them. Our focused attention module leads to state-of-the-art recovery of relations in a relationship proposal task and boosts performance for various vision and language tasks.

\end{abstract}

%% file: tex/introduction.tex
\section{Introduction}
Complex tasks involving visual perception or language interpretation are inherently contextual. In an image of an office scene, for example, a computer mouse may be too small to detect but the recognition of a computer keyboard might hint at its presence and its possible locations. The study of objects in their context is a cornerstone of much past computer vision work \cite{belongie2007}. Scene categories are often determined by the relationships between objects or environments commonly found in them \cite{zhou2017places} while in natural language processing words must be interpreted in relation to other words or phrases in sentences. Machine learning algorithms that learn object to object or word to word relationships have thus been sought. Among them, attention networks have shown great promise for the task of learning relationship attention weights between entities \cite{velivckovic2017graph,vaswani2017attention}. As a recent example, the scaled dot product attention module from \cite{vaswani2017attention} achieves state of the art performance in language translation tasks.


We here propose to explicitly supervise the learning of attention weights between elements of a data source using a novel center-mass cross entropy loss. The minimization of this loss increases relation weights between entity pairs which are more commonly observed in the data, but without the need for handcrafted frequency measurements. 
We design
a focused attention network that is end-to-end trainable and which explicitly learns pairwise element affinities without the need for relationship annotations in the data. Multiple experiments demonstrate that such focused attention improves upon the baseline as well as attention without focus, for both computer vision and natural language processing tasks. 
In a relationship proposal task the use of this backbone achieves results comparable to the present state-of-the-art \cite{relproposal}, even without the use of ground truth relationship labels. The use of ground truth labels for focused attention learning leads to a further 25\% relative improvement, as measured by a relationship recall metric.

%% file: tex/task_limit.tex
\section{Motivation}

\paragraph{Attention Networks -- The Present State} 
The modeling of relations between objects as well as objects in their common contexts has a rich history in computer vision \cite{belongie2007, torralba2003contextual, galleguillos2010context}. Deep learning based object detection systems leverage attention models to this end, to achieve impressive performance in recognition tasks. The scaled dot product attention module of \cite{vaswani2017attention}, for example, uses learned pairwise attention weights between region proposal network (RPN) generated bounding boxes in images of natural scenes \cite{hu2017relation} to boost object detection.
Pixel level attention models have also been explored to aid semantic segmentation \cite{zhao2018psanet} and video classification \cite{wang2018nonlocal}.

\begin{figure}[h]
    \centering
    \begin{tabular}{c c | c}
         \includegraphics[width=0.27\textwidth, trim={1.8cm 0cm 3cm 0cm}, clip]{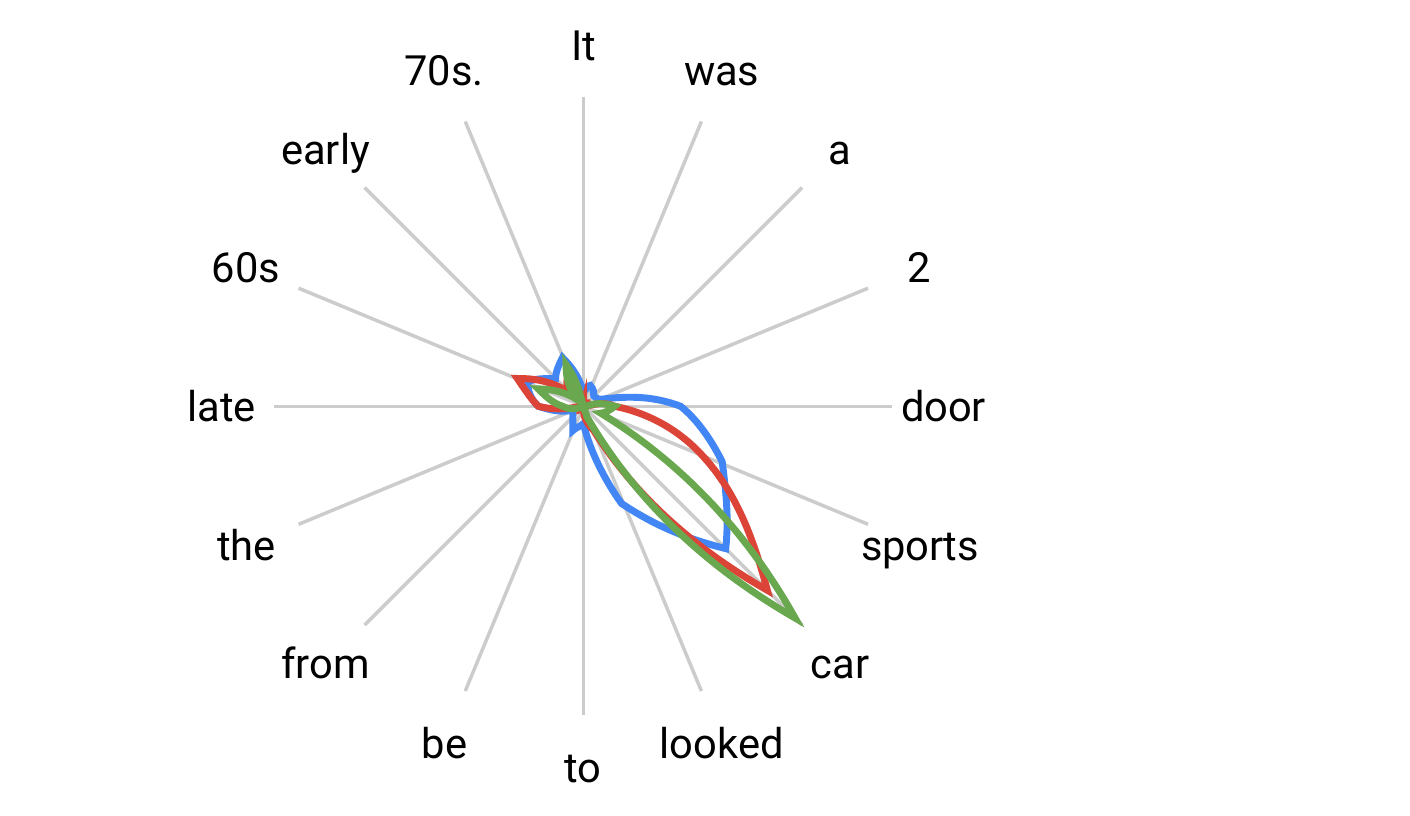}
         & 
         \includegraphics[width=0.36\textwidth, trim={1.8cm 0cm 0cm 0cm}, clip]{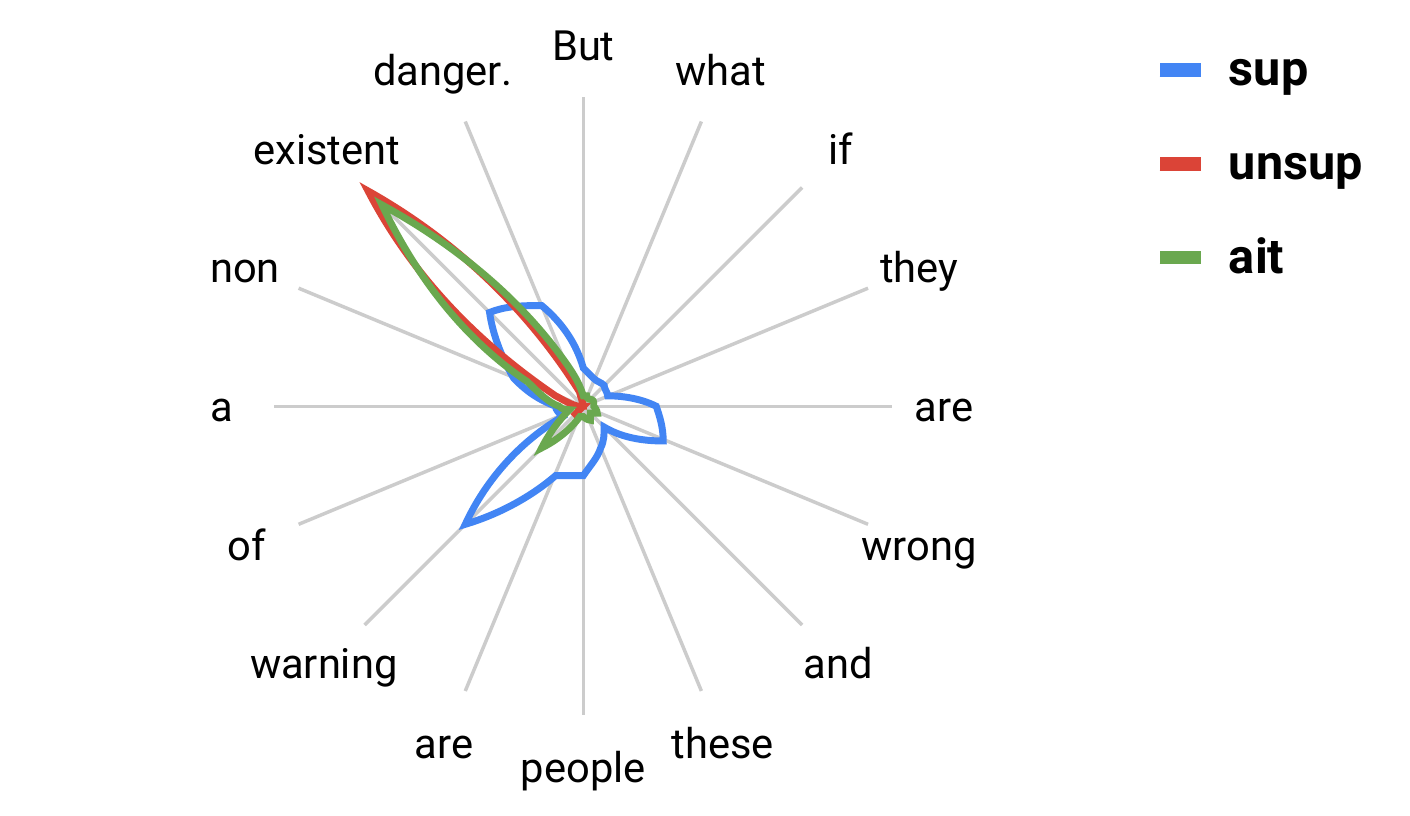}
         &
         \includegraphics[width=0.27\textwidth, trim={0cm 0cm 14cm 0cm}, clip]{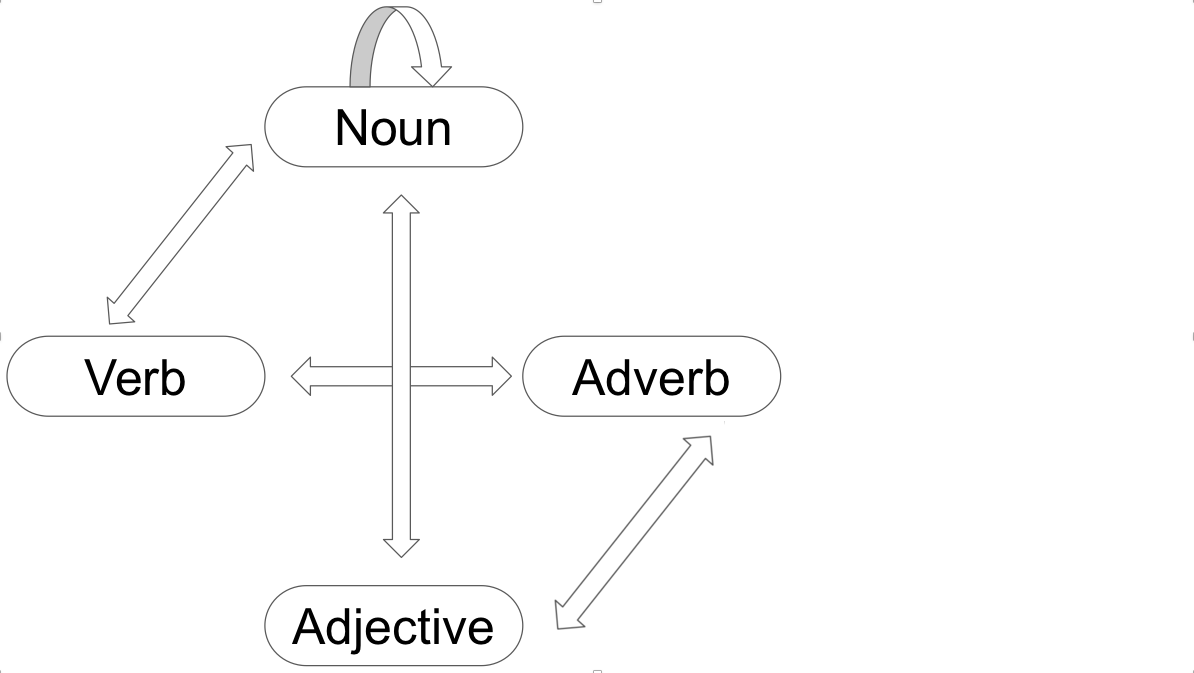} 
         \\
    \end{tabular}
    \caption{Left: Visualization of the word importance factor, which models the contribution of a given word to its sentence level context (see Section \ref{sec:nlp_task}). The sentence is in a clockwise direction starting from 12 o'clock. "ait": weights learned using Hierarchical Attention Networks \cite{hatt}, "unsup": weights from the unsupervised case of our Focused Attention Network module, and "sup": weights from the supervised case. Right: semantic word-to-word relationship labels used in the supervision of our network (see Section \ref{sec:suptarget}).}
    \label{fig:rel_nlp_comp}
\end{figure}

Current approaches to learn the attention weights 
often do not reflect relations {\em between} entities in a typical visual scene. In fact, for a given reference object (region), relation networks \cite{hu2017relation} tend to predict high attention weights with scaled or shifted bounding boxes surrounding the same object instance. This is likely because including surrounding context, or simply restoring missing parts of the reference object, boosts object detection. The learned relationship weights between distinct objects (regions) are also often small in magnitude. Typical qualitative examples comparing Relation Networks with our Focused Attention Network are shown in Figure \ref{fig:rel_comparison}, with a quantitative comparison reported in Section \ref{sec:exp}. Similar situations can occur in applications of attention networks to natural language processing tasks. In document classification, for example, attention weights learned using Hierarchical Attention Networks \cite{hatt} tend to concentrate on a few words in a sentence, while our Focused Attention Network leads to more distributed attention, allowing for more comprehensive sentence level features, as illustrated in Figure \ref{fig:rel_nlp_comp}.

\begin{table}[t]
    \centering
    \vspace{-1cm}
    \hspace*{-0.45cm}
    \begin{tabular}{c c | c c}
    \multicolumn{2}{c|}{Entities Attended to } & \multicolumn{2}{c}{Quality of Attention Weights}  \\ 
    \hline
    Rel. Networks & Focused Attn. Networks & Rel. Networks & Focused Attn. Networks \\
    \centering
      \includegraphics[width=32mm, trim={0 1cm 0 4cm}, clip]{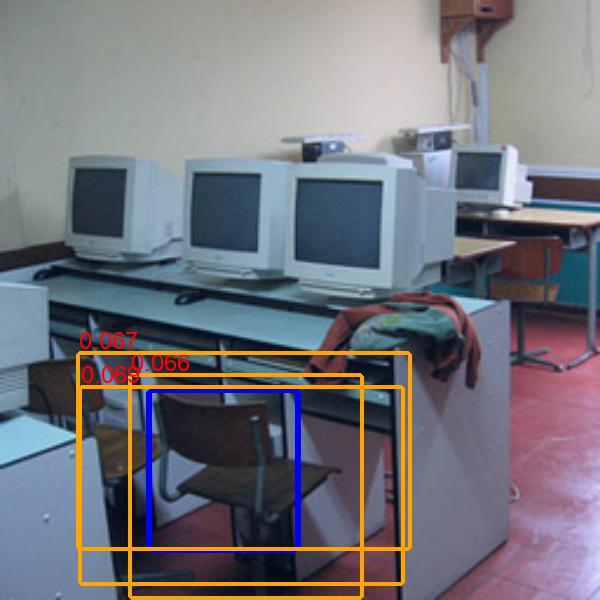} & 
      \includegraphics[width=32mm, trim={0 1cm 0 4cm}, clip]{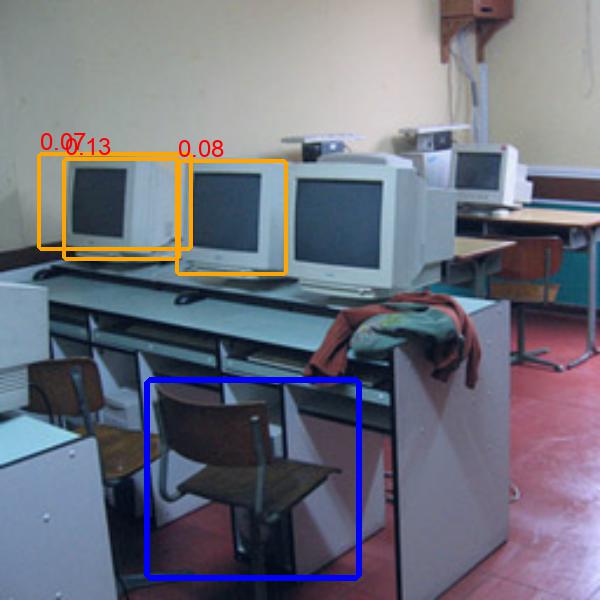} &
      \includegraphics[width=32mm]{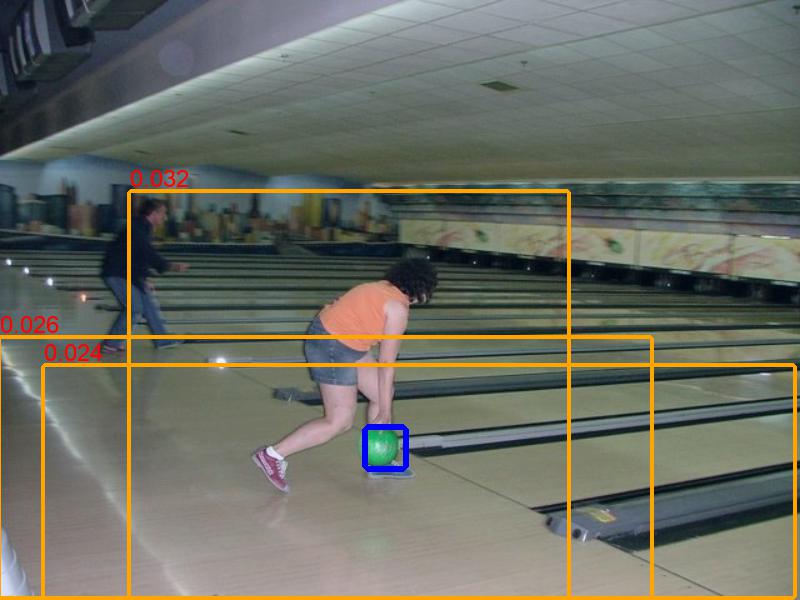} & 
      \includegraphics[width=32mm]{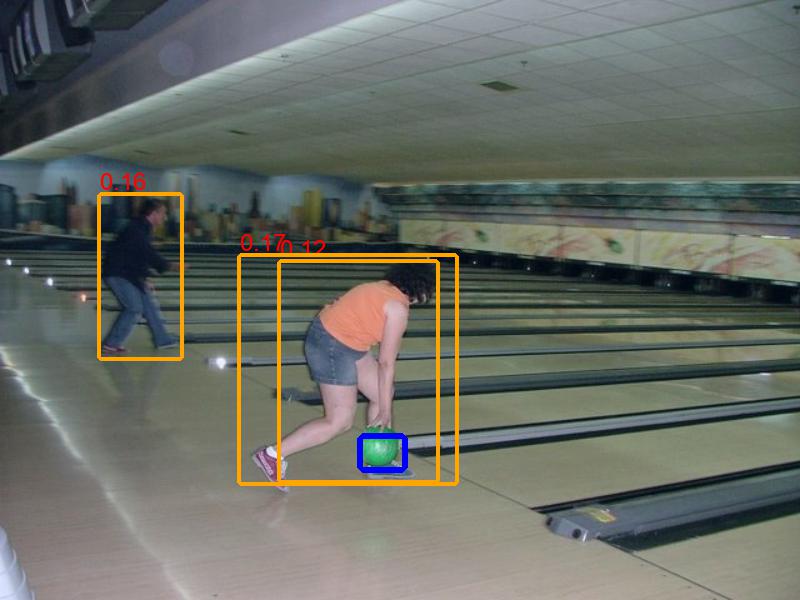} \\
    \end{tabular}
    \captionof{figure}{A comparison of recovered relationships on the MIT67 dataset, with training {\em only} on the minicoco dataset. The reference object is surrounded by a blue box and regions with which it learns relationships are shown with orange boxes with the relationship weights visible in red text (zoom-in on the PDF).
    Left: Relation Networks \cite{hu2017relation} often learn weights between a reference object and its immediate surrounding context, while Focused Attention Networks better emphasize relationships between distinct and spatially separated objects. Right: Relation Networks can suffer from a poor selection of regions to pair, or low between object relationship weights in comparison to Focused Attention Networks.  
    }
    \vspace{-0.2cm}
    \label{fig:rel_comparison}
\end{table}

\paragraph{Attention Networks -- Limitations} A present limitation of attention networks in various applications is the use of only task specific losses as their training objectives. There is little work thus far on explicitly supervising the learning of weights, so as to be more distributed across meaningful entities. 
For example, Relation Networks \cite{hu2017relation} and those applied to segmentation problems, such as PSANet \cite{zhao2018psanet}, learn attention weights {\em solely} by minimizing categorical cross entropy for classification, L1 loss for bounding box localization or pixel-wise cross entropy loss for semantic segmentation \cite{zhao2018psanet}.
In language tasks including machine translation \cite{vaswani2017attention} and document classification \cite{hatt}, the attention weights are also solely learned by minimizing the categorical cross entropy loss. In this article we refer to such attention networks as being
\textit{unsupervised}.


Whereas attention aggregation with learned weights boosts performance for these specific tasks, our earlier examples provide evidence that relationships between distinct entities may not be adequately captured. We shall address this limitation by focusing the attention weight learning using a novel center-mass cross entropy loss, as discussed in the
following section.


%% file: tex/att_sup.tex
\section{Focusing the Attention}\label{sec:att_sup}
Given our goal of better reflecting learned attention weights between distinct entities, we propose to explicitly supervise the learning of attention relationship weights. We accomplish this by introducing a novel center-mass cross entropy loss.

\subsection{Problem Statement}\label{sec:problem_state}
Given a set of $N$ entities that are generated by a feature embedding framework, which can be a region proposal network (RPN) \cite{fasterRCNN} or a word embedding layer with a bidirectional LSTM \cite{hatt}, for the $i$-th entity we define $\mathbf{f}^i$ as the embedding feature. To compute the relatedness or affinity between entity $m$ and entity $n$, we define an attention function $\mathcal{F}$ which computes the pairwise attention weight as
\begin{equation}
\omega^{mn} = \mathcal{F}(\mathbf{f}^m, \mathbf{f}^n).
\label{eq.att_fn}
\end{equation}
A specific form of this attention function applied in this paper is reviewed in Section \ref{sec:dotprodatt}, and it originates from the scaled dot product attention module of \cite{vaswani2017attention}. 

We can now build an attention graph $G$ whose vertices $m$ represent entities in a data source with features $F = \{\mathbf{f}^m\}$ and whose edge weights $\{\omega^{mn}\}$ represent pairwise affinities between the vertices. We define the graph adjacency matrix for this attention graph as $\mathcal{W}$. We propose to supervise the learning of $\mathcal{W}$ so that the matrix entries $\omega^{mn}$ corresponding to entity pairs with high co-occurrence in the training data gain higher attention weights.


\subsection{Supervision Target} \label{sec:suptarget}
We now discuss how to construct ground truth supervision labels in matrix form to supervise the learning of the entries of $\mathcal{W}$. For {\bf visual recognition} tasks we want our attention weights to focus on relationships between objects from different categories, so for each entry $t^{mn}$ of the ground truth relationship label matrix $\mathcal{T}$, we assign $t^{mn} = 1$ only when: 1) entities $m$ and $n$ overlap with two different ground truth objects' bounding boxes with intersection over union (IOU) $> 0.5$ and 2) their category labels $c_m$ and $c_n$ are different. 
We provide a visualization of such ground truth relationships in Figure \ref{fig:visual_target} of the appendix.
For {\bf language} tasks we want the attention weights to reveal meaningful word pairs. For example, semantic relationships between nouns and nouns, verbs and nouns, nouns and adjectives, adverbs and verbs, and adverbs and adjectives should be encouraged. To this end, we build a simple word category pair dictionary of valid pairings (see Figure \ref{fig:rel_nlp_comp}) and assign label $t^{mn} = 1$ when the word category pair $c_m$ and $c_n$ is found in this semantic pair dictionary. 

\paragraph{Center-Mass}
Intuitively, we would like $\mathcal{W}$ to have high affinity weights at those entries where $t^{mn} = 1$, and low affinity weights elsewhere, i.e., the attention weights should concentrate on the 1's in the ground truth relationship label matrix $\mathcal{T}$. We capture this via a notion of {\em center-mass} $\mathcal{M}$ of ground truth relation weights, defined as 
\begin{equation}
\mathcal{M} = \sum  \mathcal{\tilde W} \odot  \mathcal{T},
\label{eq.center_mass}
\end{equation}
where $\mathcal{\tilde W} = softmax( \mathcal{W} ) $ is a matrix-wise softmax operation.

\subsection{Center-mass Cross Entropy Loss} 
\label{sec:centermass}
The key to our approach is the introduction of a center-mass cross entropy loss, which aims to focus attention weight learning so that $\omega^{mn}$ is high for pairs of commonly occurring distinct entities. 
The loss is computed as 
\begin{equation}
\mathcal{L} = - (1 - \mathcal{M})^r \log (\mathcal{M}).
\label{eq.rel_loss}
\end{equation} 

When minimizing this center-mass loss 
more frequently occurring 1-labeled pairs in the matrix will cumulatively receive stronger emphasis, for example, human-horse pairs versus horse-chair pairs in natural images.
Furthermore, when supervising the attention learning in conjunction with another task specific loss, the matrix entries that reduce the task loss will also be optimized. The resultant dominant $\omega^{mn}$ entries will not only reflect entity pairs with high co-occurrence, but will also help improve the main objective. The focal term $(1- \mathcal{M})^r$ \cite{focalloss} helps narrow the gap between well converged center-masses and those that are far from convergence. For example, with a higher center-mass value the gradient on the log loss will be scaled down, whereas for a lower center-mass the gradient will be scaled up.
The focal term prevents the network from committing only to the most dominant $\omega^{mn}$ entries, and thus promotes diversity. We choose $r=2$ in our experiments, motivated by the model ablation study conducted in Section \ref{sec:exp_ablation}.


%% file: tex/net_architecture.tex
\section{Network Architecture}
Our focused attention module originates from the scaled dot product attention module in \cite{vaswani2017attention}. We discuss our network structures for the focused attention weight learning backbone and various specific tasks, as shown in Figure \ref{fig:net_architecture}, with implementation details provided in the appendix.
\begin{figure}[t]
    \centering
    \vspace{-1.5cm}
    \includegraphics[width=0.95\textwidth, trim={3cm 3.5cm 3cm 0cm}, clip]{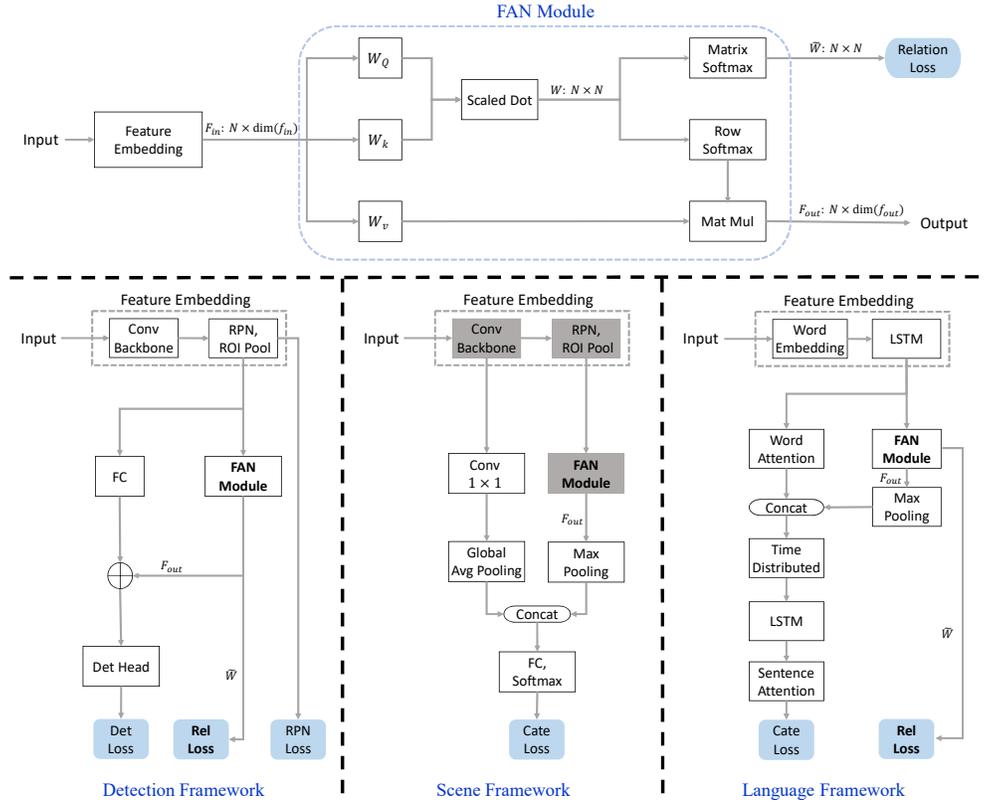}
    \caption{Top: The Focused Attention Network backbone. Bottom left: we add a detection branch to the backbone, similar to that of \cite{hu2017relation}. 
    Bottom middle: we add a scene recognition branch to the backbone. Bottom right: we insert the Focused Attention Module into a Hierarchical Attention Network \cite{hatt}. Here ``FC" stands for fully connected layer and ``Conv" stands for convolutional layer.
    }
    \label{fig:net_architecture}
\end{figure}

\subsection{Scaled Dot Product Attention Network}\label{sec:dotprodatt}
We briefly review the computation of attention weights in 
\cite{vaswani2017attention}, given a pair of nodes from the attention graph defined in Section \ref{sec:problem_state}. Let an entity node consist of its feature embedding, defined as $\mathbf{f}$. Given a reference entity node $m$, such as one of the blue boxes in Figure \ref{fig:rel_comparison}, the attention weight $\tilde{\omega}^{mn}$ indicates its affinity to a surrounding entity node $n$. It is computed using a softmax activation over the scaled dot products $\omega^{mn}$ defined as: 
\begin{equation}
\tilde{\omega}^{mn} = \frac{\exp(\omega^{mn})}{\sum_{k} \exp(\omega^{kn})}, \quad
\omega^{mn} = \frac{dot(W_K{\mathbf{f}}^{m}, W_Q \mathbf{f}^{n})}{\sqrt{d_k}}.
\label{eqn:object_relation_weight}
\end{equation}
Both $W_K$ and $W_Q$ are matrices and so this linear transformation projects the embedding features $\mathbf{f}^m$ and $\mathbf{f}^n$ into metric spaces to measure how well they match. The feature dimension after projection is $d_k$. With the above formulation, the attention graph affinity matrix is defined as $\mathcal{W} = \{{\omega}^{mn} \}$.

\subsection{Focused Attention Network (FAN) Backbone}
In Figure \ref{fig:net_architecture} (top) we illustrate the base Focused Attention Network architecture. Here the dot product attention weights $\mathcal{W}$ go through a matrix-wise softmax operation to generate the attention matrix output $\tilde{\mathcal{W}}$, that is used for the focused supervision with the center-mass cross entropy loss defined in Section \ref{sec:centermass}. We shall refer to this loss term as \textit{relation loss}. In parallel, a row-wise softmax is applied to $\mathcal{W}$ to output the coefficients $\mathcal{W}_{agg}$, which are then used for attention weighted aggregation: $\mathbf{f}_{out}^m = \sum_n \omega_{agg}^{mn} f^n$. 
The aggregated output from the FAN module is sent to a task specific loss function. The entire backbone is end-to-end trainable, with both the task loss and the relation loss. 

\subsection{Object Detection and Relationship Proposals} \label{sec:detection}
In Figure \ref{fig:net_architecture} (bottom left) we demonstrate how to generalize the FAN module for object detection and relationship proposal generation. The network is end-to-end trainable with detection loss, RPN loss and our relation loss. In addition to the ROI pooling features $\mathbf{F} \in \mathcal{R}^{N_{obj} \times 1024}$ from the Faster R-CNN backbone of \cite{fasterRCNN}, contextual features $\mathbf{F}_c$ from attention aggregation are applied to boost detection performance: $\mathbf{F}_c = \mathcal{W}_{agg} \mathbf{F}$.
The final feature descriptor for the detection head is $\mathbf{F}_d = \mathbf{F} + \mathbf{F}_c$, following \cite{hu2017relation}. In parallel, the attention matrix output $\tilde{\mathcal{W}} \in \mathcal{R}^{N \times N}$ is used to generate relationship proposals by finding the top K weighted pairs in the matrix. 


\subsection{Scene Categorization Task} \label{sec:scene_net}
In Figure \ref{fig:net_architecture} (bottom middle) we demonstrate how to apply the FAN module to scene categorization. Since there are no bounding box annotations in most scene recognition datasets, we adopt a pre-trained FAN detection module, described in Section \ref{sec:detection}, in conjunction with a newly added convolution branch, to perform scene recognition. From the convolution backbone, we apply an additional convolution layer followed by a global average pooling to acquire the scene level feature descriptor $\mathbf{F}_s$. The FAN module takes as input the object proposals' visual features $\mathbf{F}$, and outputs the aggregation result as the scene contextual feature $\mathbf{F}_c$. The input to the scene classification head thus becomes $\mathbf{F}_{meta} = concat(\mathbf{F}_s, \mathbf{F}_c)$, and the class scores are output. 




\subsection{Document Categorization Task}\label{sec:nlp_task}
In Figure \ref{fig:net_architecture} (bottom right) we demonstrate how to apply the FAN module to a document classification task, using Hierarchical Attention Networks \cite{hatt}. We insert the FAN module into the word level attention layer, but making it parallel to the original word-to-sentence attention module, and denote it as ``FAN-hatt''. FAN module learns word-to-word attention
and outputs a sentence level descriptor. 
The FAN descriptor is concatenated with the output from the word-to-sentence attention to result in enhanced sentence level embedding. The rest of the computation follows the original paper \cite{hatt}, where a sentence level attention model is applied and a final document level descriptor is abstracted.

%% file: tex/exp.tex
\section{Experiments} \label{sec:exp}
\subsection{Datasets and Network Training} \label{sec:data_train}
\textbf{Datasets.} We evaluate our FAN architecture using the following datasets: VOC07, MSCOCO, Visual Genome, MIT67, 20 Newsgroups, Yahoo Answers. Details are provided in the appendix.\\
\textbf{Vision Network Training} 
Following Section \ref{sec:detection}, we first train the detection-and-relation joint framework end-to-end with a detection task loss and a relation loss on the minicoco dataset, dubbing this network ``FAN-minicoco''. We report detection results as well as relation learning quality. We then fine tune the scene task structure on the MIT67 dataset, using the pre-trained FAN-minicoco network (see Section \ref{sec:scene_net}), and report scene categorization performance. Details on the network architecture, the input/output dimensions and the hyper-parameters are in the appendix.\\
\textbf{Language Network Training}
We train the FAN models with Hatt \cite{hatt} as the backbone. Details on the architecture and hyper-parameters are in the appendix.

\subsection{Evaluation Metrics}
\label{sec:evaluation}
\textbf{Object Detection Task.} For the VOC07 and MSCOCO datasets the evaluation metric is mAP (mean average precision), as defined in \cite{voc, mscoco}.\\ 
\textbf{Relationship Proposal Task.} We evaluate the learned relationships using a recall metric which measures the percentage of ground truth relations that are covered in the predicted top K relationship list, which is consistent with \cite{relproposal,neuralmotifs,xu2017scenegraph}. A detailed definition is in the appendix.\\
\textbf{Classification Tasks.} For MIT67, 20 Newsgroups and Yahoo Answers, where the task is to classify scenes or documents, we use classification accuracy as the evaluation metric.


\subsection{Model Ablation Study} \label{sec:exp_ablation}
Prior to evaluating the FAN model we carry out ablation studies to examine ways of supervising the center mass and different choices of loss functions.
\paragraph{Focused Supervision Strategies}
We consider different approaches to training the focused attention, using the detection-and-relation joint framework from Section \ref{sec:detection} on the VOC07 dataset for each case. First, we apply a row-wise softmax over the pre-activation matrix $\mathcal{W}$ and calculate the center-mass in a row-wise manner and apply the center-mass cross entropy loss accordingly. We refer to this as ``row''. Second, we apply the supervision explained in Section \ref{sec:centermass} but without the use of the focal term, and refer to this as ``mat''. Third, we add the focal term to the matrix supervision,  referring to this as ``mat-focal''. The results in Table \ref{tab:sup_eval} show that the focused attention weights, when supervised using the center-mass cross entropy loss with a focal term (Section \ref{sec:centermass}), are better concentrated on inter-object relationships, as reflected by the recall metric (Section \ref{sec:evaluation}) when compared with the unsupervised case. 
In all further experiments, unless stated otherwise, we apply the matrix supervision with the focal term, since it provides the best performance.
 
\begin{table}[h]
\centering
    \begin{minipage}[t]{.475\textwidth}
        \centering
        \begin{tabular}{c}
             \includegraphics[width=0.95\textwidth, trim={2cm 7cm 0cm 8.5cm}, clip]{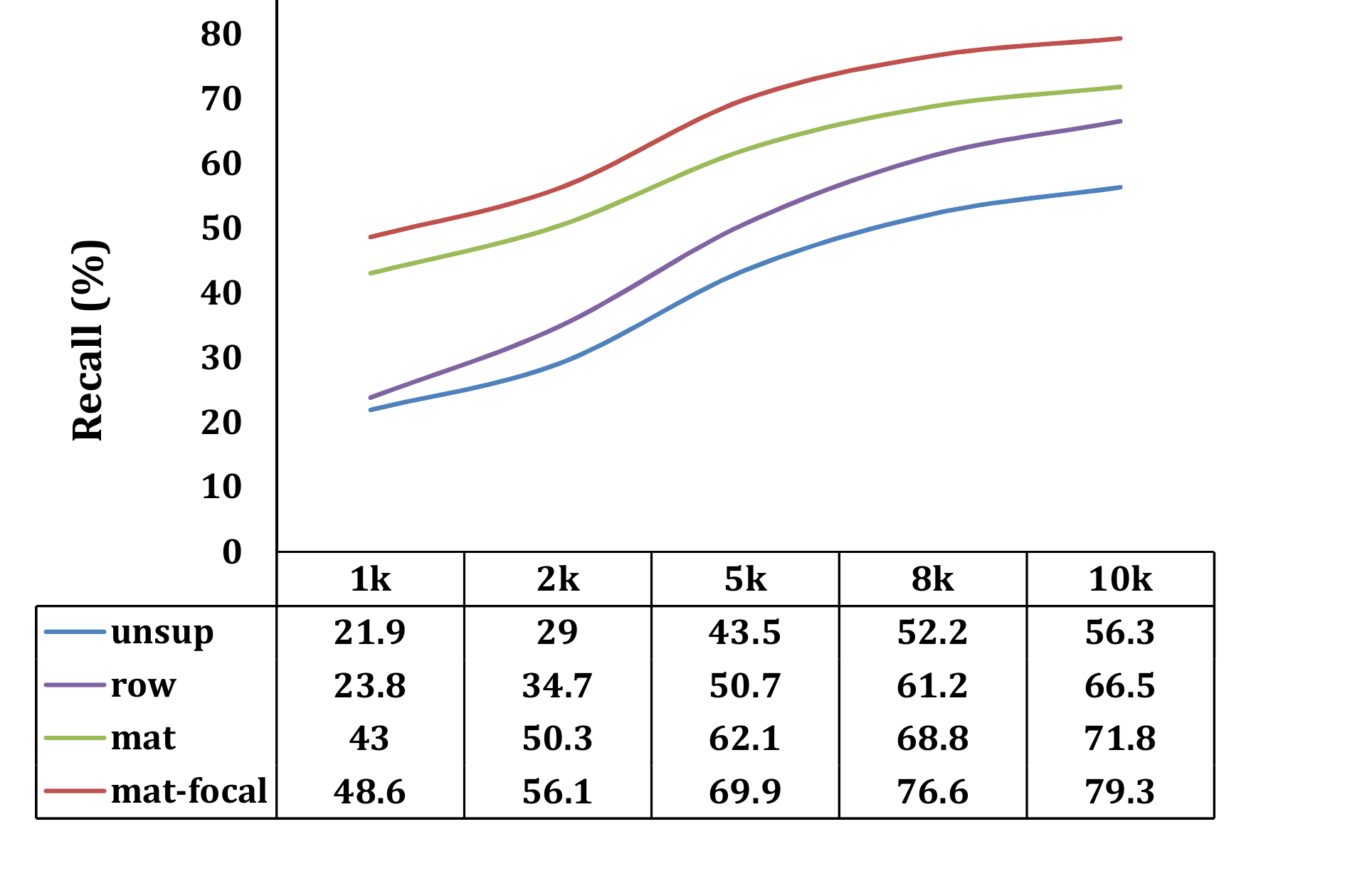}\\
        \end{tabular}
        \caption{Evaluating different supervision strategies with varying top K, using the VOC07 testset, with ground truth relation labels as described in Section \ref{sec:suptarget}.}
        \label{tab:sup_eval}
    \end{minipage}
    \hspace{0.25cm}
    \begin{minipage}[t]{.475\textwidth}
        \centering
        \begin{tabular}{c}
             \includegraphics[width=0.95\textwidth, trim={2cm 7cm 0cm 8.5cm}, clip]{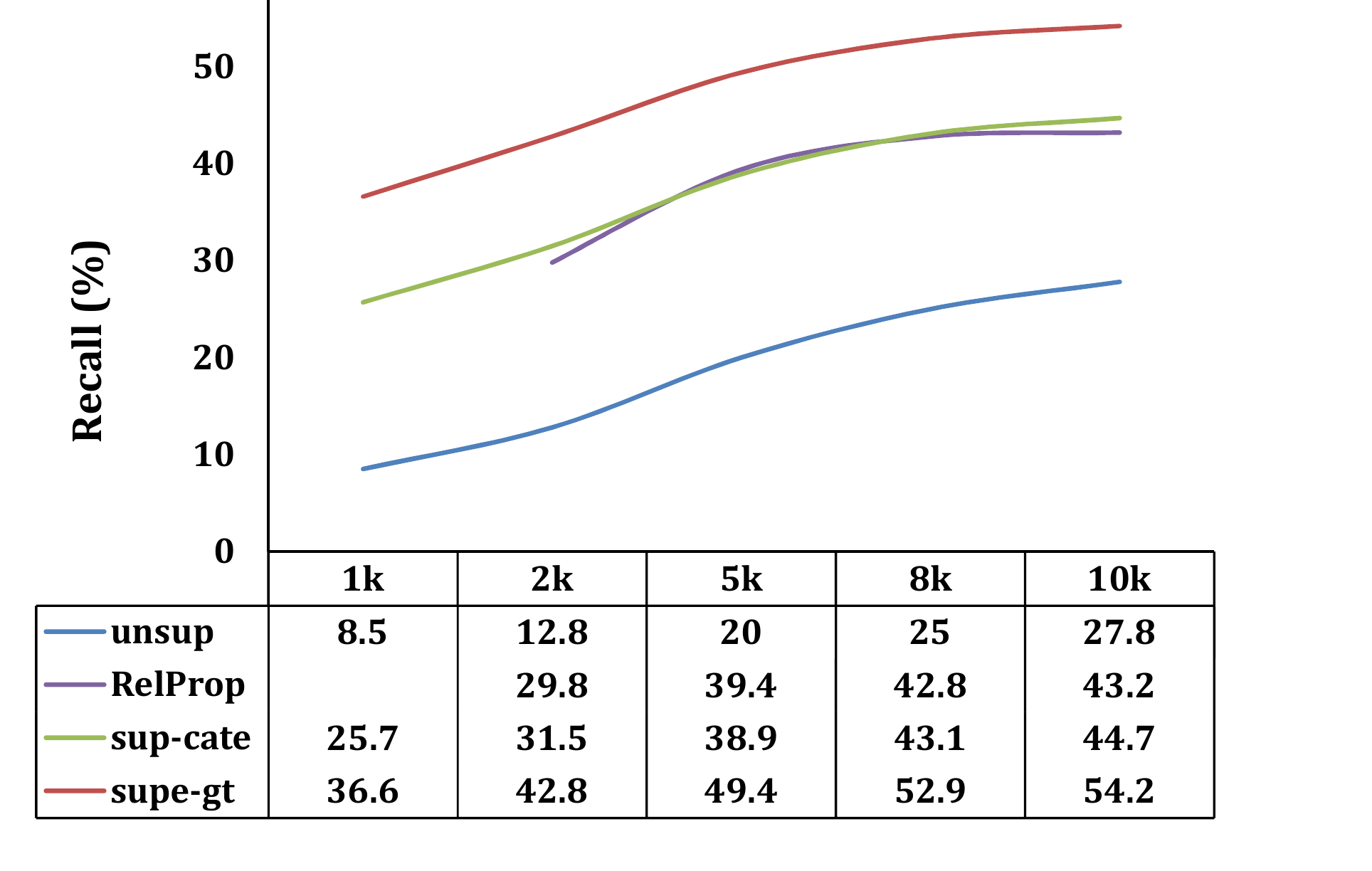}
        \end{tabular}
        \caption{Recall comparison for the Visual Genome dataset with varying top K, where the ground truth relation labels are human annotated. See text in Section \ref{sec:relationshipproposal} for a discussion. }
    \label{tab:rel_eval}
    \end{minipage}
\end{table}

\paragraph{Design of Loss Functions.} 
\begin{table}[h]
	\centering
	\small 
	\begin{tabular}{@{}l|c|c||c|c|c|c|c}
		\hline
		VOC07        & smooth L1        & L2             & $r=$ 0   &  $r=$ 1   &           $r=$ 2   &  $r=$ 3    &  $r=$ 4 \\ \hline
		mAP@0.5 &   79.6 $\pm$ 0.2 & 79.7 $\pm$ 0.2 & 79.4 $\pm$ 0.1 & 79.5 $\pm$ 0.2 & \textbf{79.9 $\pm$ 0.2} & 79.8 $\pm$ 0.1  & 79.7 $\pm$ 0.2  \\
		recall@5k & 60.3 $\pm$ 0.3 & 64.6 $\pm$ 0.5 & 62.1 $\pm$ 0.3 & 66.5 $\pm$ 0.2 & \textbf{69.9 $\pm$ 0.3} & 68.7 $\pm$ 0.4  & 67.1 $\pm$ 0.3  \\
		\hline
	\end{tabular}
	\caption{An evaluation of smooth $L_1$ and $L_2$ loss functions and variations of the focal loss factor $r$, on the VOC07 dataset. The results are reported as percentages (\%) averaged over 3 runs. 
	}
	\label{tab:loss_ablation}
\end{table}
We conduct a second ablation study to examine additional loss functions for optimizing the center mass as well as varying focal terms $r$, as introduced in Section \ref{sec:centermass}. Defining $x =  1 - \mathcal{M} \in [0, 1]$, we consider loss function variants $L_2 = x^2$ and
\begin{equation}
  \textrm{smooth}_{L_1}(x) =
  \begin{cases}
    x^2 \quad \textrm{if } \quad |x| < 0.5\\
    |x| - 0.25 \quad \textrm{otherwise},
  \end{cases}
\end{equation}
in addition to the focal loss. The results in Table \ref{tab:loss_ablation} show that focal loss is in general better than smooth L1 and L2 when supervising the center mass. In subsequent experiments we apply focal loss with $r=2$, which empirically gives the best performance.

\subsection{Relationship Proposal Task}
\label{sec:relationshipproposal} Table \ref{tab:rel_eval} shows the evaluation of relationships learned on the Visual Genome dataset, by the unsupervised Focused Attention Network model ``unsup'' (similar to \cite{hu2017relation}), the Focused Attention Network supervised with weak relation labels described in Section \ref{sec:suptarget} ``sup-cate'', and supervision with human annotated ground truth relation labels ``sup-gt''. We also include the reported recall metric from Relationship Proposal Networks \cite{relproposal}, which is a state-of-the-art level relationship learning network with strong supervision, using ground truth relationships. 
Our center-mass cross entropy loss does not require potentially costly human annotated relationship labels for learning, yet it achieves the same level of performance as the present state-of-the-art \cite{relproposal} (the green curve in Table \ref{tab:rel_eval}). When supervised with the ground truth relation labels instead of the weak labels (Section \ref{sec:suptarget}), we significantly outperform relation proposal networks (by about 25\% in relative terms for all K thresholds) with this recall metric (the red curve in Table \ref{tab:rel_eval}).

\subsection{Object Detection Task}
In Table \ref{tab:objectdetection} we provide object detection results on the VOC07 and MSCOCO datasets. In both cases we improve upon the baseline and slightly outperform the unsupervised case (similar to Relation Networks \cite{hu2017relation}). This suggests that relation weights learned using our focused attention network are at least as good as those from \cite{hu2017relation}, in terms of object detection performance.
\begin{table}[!h]
	\centering
	\small
	\begin{tabular}{@{}l|c|c|c}
		\hline
		VOC07 & base F-RCNN & FAN + $\mathcal{L}_{det}$  \cite{hu2017relation} & FAN + $\mathcal{L}_{det}$  + $\mathcal{L}_{rel}$ \\ \hline
		avg mAP (\%) & 47.0  & 47.7 $\pm$ 0.1  & 48.2 $\pm$ 0.2   \\
		mAP@0.5 (\%) & 78.2  & 79.3 $\pm$ 0.2  & 79.9 $\pm$ 0.2   \\
		\hline
		\hline
		mini COCO & base F-RCNN & FAN + $\mathcal{L}_{det}$ \cite{hu2017relation} & FAN + $\mathcal{L}_{det}$  + $\mathcal{L}_{rel}$ \\ \hline
		avg mAP (\%) & 26.8  & 27.5  & 27.9   \\
		mAP@0.5 (\%) & 46.6  & 47.4  & 47.8   \\
		\hline
	\end{tabular}
	\caption{Object Detection Results. mAP@0.5: average precision over a bounding box overlap threshold as $IOU=0.5$. avg mAP: averaged mAP over multiple bounding box overlap thresholds. VOC07 experiments are reported over 3 runs, demonstrating stability. $\mathcal{L}_{det}$ stands for detection task loss as defined in \cite{fasterRCNN} and $\mathcal{L}_{rel}$ for the center mass relation loss defined in section \ref{sec:centermass}.}
	\label{tab:objectdetection}
\end{table}

\subsection{Scene Categorization Task}
\label{sec:scenecategorization}

\begin{table}[h]
	\centering
	\small
	\begin{tabular}{@{}l|c|c|c|c|c}
		\hline
		~ & CNN & CNN & CNN + ROIs & CNN + FAN-unsup & CNN + FAN-sup\\ \hline
		Pretraining  & Imgnet  & Imgnet+COCO  & Imgnet+COCO & Imgnet+COCO & Imgnet+COCO   \\ \hline
		Features & $\mathbf{F}_S$  & $\mathbf{F}_S$  & $\mathbf{F}_S, max(\mathbf{F})$ & $\mathbf{F}_S, \mathbf{F}_C$  & $\mathbf{F}_S, \mathbf{F}_C$\\ \hline
		Accuracy (\%) & 75.1  & 76.8  & 78.0 $\pm$ 0.3 & 77.1 $\pm$ 0.2 & 80.2 $\pm$ 0.3   \\
		\hline
	\end{tabular}
	\caption{MIT67 Scene Categorization Results. The important entries are averages over 3 runs. 
	See the text in Section \ref{sec:scenecategorization} for a discussion.
	For details regarding $F_s$, $F_c$ and $F$, see Section \ref{sec:scene_net}.}
	\label{tab:scenecategorization}
\end{table}

We adopt the FAN-minicoco network (Section \ref{sec:data_train}), and add an additional scene task branch to fine tune it on MIT67, as discussed in Section \ref{sec:scene_net}. Table \ref{tab:scenecategorization} shows the results of applying this model to the MIT67 dataset.
We refer to the backbone as ``CNN'' (first column), which is the left branch in Figure \ref{fig:net_architecture} (bottom middle). In the second column we apply the same network further fine-tuned on minicoco before training on MIT67. In the third column we include the detection branch, which is the right branch in Figure \ref{fig:net_architecture} (bottom middle) but remove its FAN module.
In the fourth and fifth columns we apply the full scene architecture in Figure \ref{fig:net_architecture} (bottom middle), using FAN-minicoco network pretrained without (unsupervised) and with (supervised) relation loss, respectively. The FAN supervised case (fifth column) demonstrates a non-trivial improvement over the baseline (third column) and also significantly outperforms the unsupervised case (fourth column). This suggests that the relation weights learned solely by minimizing detection loss do not generalize well to a scene task, whereas those learned by our Focused Attention Network supervised by weak relations labels can. We hypothesize that recovering informative relations between distinct objects, which is what our Focused Attention Network is designed to do, is particularly beneficial for scene categorization.

\subsection{Document Categorization Task}

\begin{table}[!h]
    \centering
    \small
    \begin{tabular}{lc ||ccc}
        \toprule
         Datasets & Train Vol. Per Cate & Hatt \cite{hatt} & FAN-hatt-unsup  & FAN-hatt-sup    \\

        \midrule
         20 News & 550& 64.0 $\pm$ 0.3 & 64.6 $\pm$ 0.2 & 65.6 $\pm$ 0.2 \\
         Yahoo-mini & 5,000 & 64.9 $\pm$ 0.2 & 65.2 $\pm$ 0.2 & 66.0 $\pm$ 0.1  \\
         \hline
         \hline
         Yahoo-half & 70,000 & 72.2 $\pm$ 0.1 & 72.1 $\pm$ 0.1 & 72.4 $\pm$ 0.1  \\
        \bottomrule
         
    \end{tabular}
    \caption{Document categorization results for the 20 Newsgroups and Yahoo Answers datasets, with the results averaged over 5 trials. For the Yahoo dataset, we train on sub-sampled training sets and report results on the full test set.}
    \label{tab:20news}
\end{table}

We present document classification results in Table \ref{tab:20news}. We provide a comparison between the base Hierarchical Attention Networks (Hatt), and FAN-hatt, explained in Section \ref{sec:nlp_task}, with and without relation loss supervision. 
The supervised (semantic) focused attention supervision results in an improvement over both the unsupervised case and the baseline, particularly when the training data per category is small (top two rows). This advantage diminishes, however, when the training volume is dramatically increased, suggesting that in the latter case a baseline network is able to perform equally well. In addition, the qualitative distributions of word importance factors (see appendix for an exact definition) in Figure \ref{fig:rel_nlp_comp} suggest that focused semantic attention encourages more diversity in attention weight learning, which in turn leads to better document classification performance. 
 

%% file: tex/conclusion.tex
\section{Conclusion}
\label{sec:conclusion}
 
 Our Focused Attention Network is versatile, and allows the user to direct the learning of attention weights in the manner they choose. The application of the FAN module allows multiple informative entities in the data source to contribute to the learned feature representation, abstracting diverse aspects of the data.  In multiple experiments we have demonstrated the benefit of learning relations between distinct objects for computer vision tasks, and between lexical categories (words) for document classification tasks. It not only boosts performance in object detection, scene categorization and document classification, but also leads to state-of-the-art performance in a relationship proposal task. In the future we envision its use as a component for deep learning architectures where supervised control of relationship weights is desired, since it is adaptable, modular, and end-to-end trainable in conjunction with a task specific loss.

%% file: tex/supplementary.tex
\section{Further Experimental Details}
\subsection{Datasets}

\textbf{VOC07:} is part of the PASCAL VOC detection dataset \cite{voc}. It consists of 5k images for training and 5k for testing. We used the full training/test split in our experiments. \\ 
\textbf{MSCOCO:} consists of 80 object categories \cite{mscoco}. Within the 35k validation images of the COCO2014 detection benchmark, a selected 5k subset named ``minival'' is commonly used when reporting test time performance, for ablation studies \cite{hu2017relation}. We used the 30k validation images for training and the 5k ``minival'' images for testing. We define this split as ``minicoco''. \\
\textbf{Visual Genome:} is a large scale relationship understanding benchmark \cite{visualgenome}, consisting of 150 object categories and human annotated relationship labels between objects. We used 70k images for training and 30K for testing, as in the scene graph literature \cite{neuralmotifs,xu2017scenegraph}.\\ 
\textbf{MIT67:} is a scene categorization benchmark which consists of 67 scene categories, with each category having 80 training images and 20 test images \cite{mit67}. We used the full training/testing split in our experiments. \\
\textbf{20 Newsgroups:} is a document classification dataset consisting of text documents from 20 categories \cite{20news}, with 11314 training ones and 7532 test ones. We used the full training/testing split in our experiments. \\
\textbf{Yahoo Answers:} is a topic classification dataset consisting of 10 categories and is organized from the Yahoo! Answers Comprehensive Questions and Answers version 1.0 dataset. Each class contains 140,000 training samples and 6,000 testing samples. In our experiments, we evaluate on the full test set but sub-sample the training set during training.
First, we randomly sample 5,000 training samples per category within the 140,000 and refer to this configuration as ``Yahoo-mini''. Next we randomly sample 70,000 training samples per category within the 140,000 and refer to this configuration as ``Yahoo-half''. We report full test set results using the above different training sets.


\subsection{Relationship Recall Metric}
We evaluate the learned relationships using a recall metric defined as  
$R_{rel} = \frac{C(rel|topK)}{C(rel)}$.
Here $C(rel)$ stands for the number of unique ground truth relations in a given image and $C(rel|topK)$ stands for the number of unique matched ground truth relations in the top-K ranked relation weight list. In the calculation of $C(rel|topK)$, we only consider a match when both bounding boxes in a given relationship pair have overlaps of more than $0.5$, with the corresponding ground truth boxes in a ground truth relationship pair. Therefore, $R_{rel}$ measures how well the top-K ranked relation weights capture the ground truth labeled relationships. 

\subsection{Word Importance Factor}
The word importance factor models the contribution of a given word to its sentence level representation. In Hierarchical Attention Networks \cite{hatt}, the word-to-sentence attention module directly models the word importance given a learned sentence representation. 
Whereas it is different from the FAN module, which models word-to-word attention, the resultant attention weights can be comparable at the sentence level. 
To this end, we define the word importance factor as $\beta_i = \sum_j \tilde{\omega}^{ji}$, because it represents the contribution of the $i$-th word in the final aggregation result $\mathcal{W}_{agg} \mathbf{F}$. A visualization of the distribution of learned word importance factors, for Hierarchical Attention Networks and Focused Attention Networks, is provided in Figure \ref{fig:rel_nlp_comp} in the main text and Figure \ref{fig:add_nlp} in this appendix. 

\subsection{Supervision Targets $\mathcal{T}$}
We consider different possibilities for supervision targets $\mathcal{T}$, as defined in Section \ref{sec:suptarget}. 

\paragraph{Vision tasks.} In our paper we focused attention between objects from different categories, and we refer to this as \textbf{different category focused supervision}. We now consider the case that attention between different object instances is focused. That is, as long as object proposal $a$ and object proposal $b$ are different object instances in an image, we consider a possible relationship between them and assign $t^{ab} = \mathcal{T}[a,b] = 1$. We refer to this as \textbf{different instance focused supervision}. 
We provide a visual example of the above mentioned supervision targets in Figure \ref{fig:visual_target}. 

\begin{table}[h]
    \centering
    \begin{tabular}{c | c}
    Different Category & Different Instance \\
    \hline
    \centering
      \includegraphics[width=0.45\textwidth, trim={0 0cm 0 0cm}, clip]{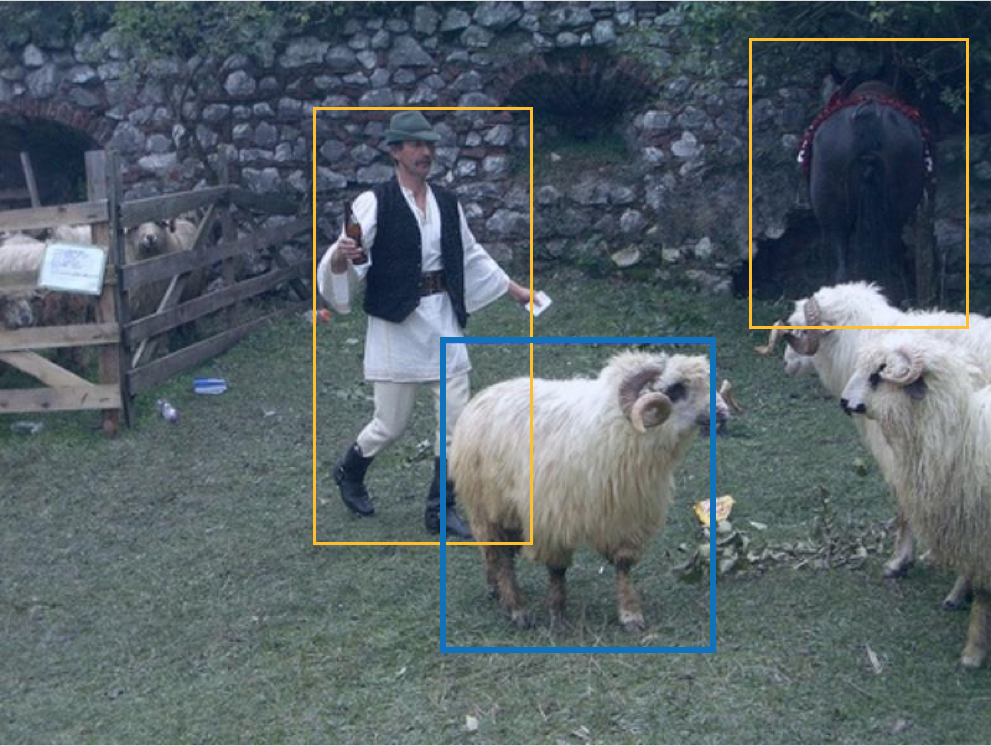} & 
      \includegraphics[width=0.45\textwidth, trim={0 0cm 0 0cm}, clip]{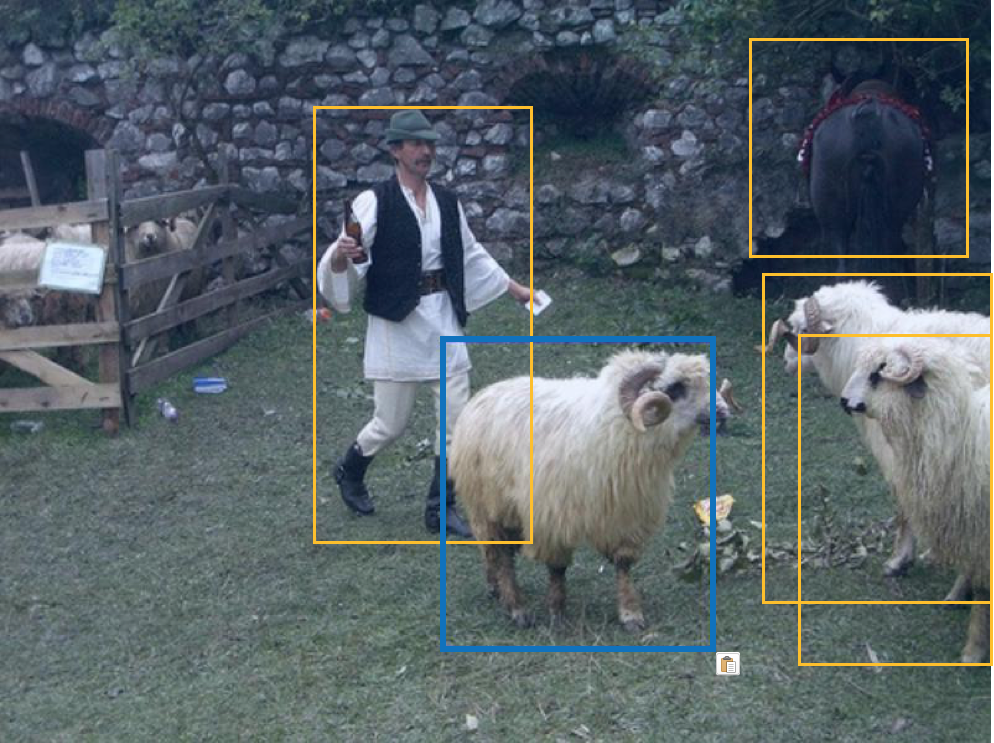} \\
    \end{tabular}
    \captionof{figure}{The visualization of supervision targets for vision tasks. The blue box indicates a fixed reference object $a$ and the orange boxes indicate the objects $b$ that have ground truth relationship with $a$, for which we assign $\mathcal{T}[a,b] = 1$. Left: different category supervision. Note that the sheep in the blue box is \textit{not} related to the other sheep in the image. Right: different instance supervision. The sheep in the blue box now has a relationship to other sheep (in yellow boxes). 
    }
    \label{fig:visual_target}
\end{table}

In Table \ref{tab:visual_sup_target}, we provide object detection results on VOC07 when training the  FAN model using different supervision targets.  
\begin{table}[!h]
    \centering
    \small
    \begin{tabular}{l|cc}
        \toprule
         VOC07 varying $\mathcal{T}$ & Diff Instance  & Diff Category    \\
        \hline
         avg mAP (\%) & 47.6 $\pm$ 0.1 & 48.2 $\pm$ 0.2\\
         mAP@0.5 (\%) & 79.5 $\pm$ 0.2 & 79.9 $\pm$ 0.2\\
        \bottomrule
    \end{tabular}
    \caption{Detection results on the VOC07 dataset when varying supervision targets, where we show maen accuracies over 3 runs.}
    \label{tab:visual_sup_target}
\end{table}

\paragraph{Language Tasks.} In our paper we focused attention between word categories according to English grammar. Here we consider additional ways of supervising the word-to-word relationships. In \textbf{different category supervision}, we consider a word pair $(a,b)$ to have a valid ground truth relationship when $a$ and $b$ belongs to different lexical categories. In \textbf{same category supervision}, we consider a word pair $(a,b)$ to have a valid ground truth relationship when $a$ and $b$ belong to the same lexical categories. In \textbf{different word supervision}, we consider a word pair $(a,b)$ to have a valid ground truth relationship when $a$ and $b$ are different words.  
We provide a comparison of using the aforementioned supervision targets on the 20News dataset, in Table \ref{tab:nlp_sup_target}, with the supervision constraints becoming stricter from the leftmost column to the rightmost column.

\begin{table}[!h]
    \centering
    \small
    \begin{tabular}{l|cccc}
        \toprule
         20news varying $\mathcal{T}$ & Diff Word & Same Category & Diff Category & Semantic    \\
        \hline
         Accuracy (\%) & 64.7 $\pm$ 0.2 & 64.9 $\pm$ 0.2 & 65.1 $\pm$ 0.1 & 65.6 $\pm$ 0.2 \\
        \bottomrule
         
    \end{tabular}
    \caption{Document categorization results for the 20 Newsgroups dataset when varying the supervision target, where we show mean accuracies over 5 runs.}
    \label{tab:nlp_sup_target}
\end{table}

\subsection{Convergence of Center-Mass}
\begin{table}[h]
    \centering
    \begin{tabular}{lcc}
        \toprule
        COCO $\mathcal{M}$ &  Training   &  Testing  \\
        \midrule
        un-sup \cite{hu2017relation} & 0.020 &  0.013   \\
        sup-obj  & 0.747 &  0.459   \\
        \bottomrule
    \end{tabular}
    \caption{We compare center-mass values for the FAN-minicoco network between training and testing. The values reported are evaluated on the minicoco train/test set.}
    \label{tab:coco_cemass}
\end{table}

We provide additional results illustrating the convergence of center-mass $\mathcal{M}$ training in Table \ref{tab:coco_cemass}. The center mass is a element wise multiplication between the post softmax attention weight matrix $ \mathcal{\tilde W}$ and the ground truth label matrix $\mathcal{T}$, defined as $ \mathcal{M} = \mathcal{\tilde W} \odot \mathcal{T} \in \left[ 0, 1\right]$. More details are in Section 3.2 of the main article.
Applying the FAN-minicoco network described in Section 5 of the main article, ``sup-obj'' stands for focusing the attention using the ground truth label constructed following Section 3.2 in the main article, and ``un-sup'' stands for the unfocused case of removing the relation loss during training, which is similar to \cite{hu2017relation}. The converged center-mass value for the supervised case is much higher than that for the unsupervised case. Empirically this suggests that our relation loss, which is designed to increase the center-mass during learning, is effective. Furthermore, the gap between the training center-mass and the testing one is reasonable for the supervised case, i.e., we do not appear to be suffering from over-fitting. We have observed the same general trends for the other tasks as well.

\section{Network Training Details}
\begin{figure}[h!]
    \centering
    \includegraphics[width=\textwidth, trim={3cm 3.5cm 3cm 6cm}, clip]{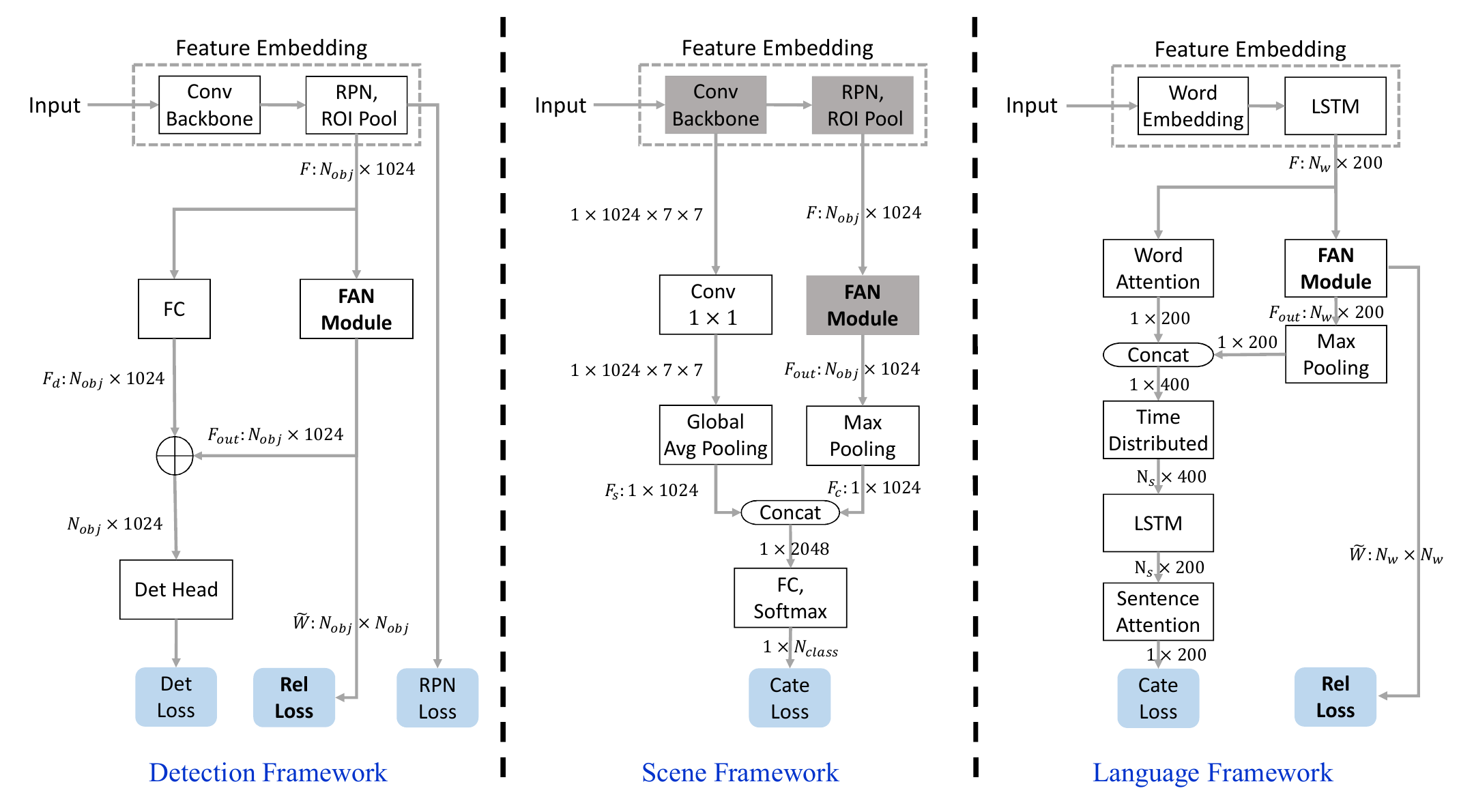}
    \caption{The input/output dimension details related to Figure 3 of the main article. The dimensions shown are for the case of a batch size of 1.
    Left: We add a detection branch to the backbone. 
    Middle: We add a scene recognition branch to the backbone.
    Right: We insert the Focused Attention Module into a Hierarchical Attention Network. 
    }
    \label{fig:detail_struct}
\end{figure}

\paragraph{Vision Tasks} Unless stated otherwise, all the vision task networks are based on a ResNet101 \cite{he2016deep} structure trained with a batch size of 2 (images), using a learning rate of $5e-4$ which is decreased to $5e-5$ after 5 epochs, with 8 epochs in total for the each training session. SGD with a momentum optimizer is applied with the momentum set as $0.9$. The number of RPN proposals is fixed at $N_{obj} = 300$. Thus the attention weight matrix $\mathcal{W}$ has a dimension of $300 \times 300$ for a single image. Further details regarding input/output dimensions of the intermediate layers can be found in Figure \ref{fig:detail_struct} (left and middle). 

\paragraph{Language Task}
For the document classification task, the network structure is based on a Hierarchical Attention Network \cite{hatt}. For all experiments, the batch size is set to be $256$ (documents), and the word embedding dimension is set to $100$. The maximum number of words in a sentence is set to be $N_w = 30$, and the maximum number of sentences in a document is set to be $N_s = 15$.  Therefore, the word level Focused Attention Network's attention weight matrix $\mathcal{W}$ has a dimension of $30 \times 30$, for a single sentence. The output dimension for Bi-LSTMs is set to be 100, and the attention dimension in attention models is also set to be 100. The Adam optimizer \cite{kingma2014adam} is applied with an initial learning rate of $1e-3$. The network is trained end-to-end with categorization loss and relation loss for 15 epochs. Further details regarding input/output dimensions of intermediate layers can be found in Figure \ref{fig:detail_struct} (right).

\section{Scaling of loss terms}
We applied a balancing factor $\lambda$ to properly weight the relation loss when training in conjunction with the main objective loss. 

\paragraph{Detection and relation proposal} We applied 
\begin{equation}
    \mathcal{L} = \mathcal{L}_{det} + \lambda \mathcal{L}_{rel}, \lambda = 0.01
\end{equation}
where $\mathcal{L}_{det}$ is defined in \cite{fasterRCNN}, which is a combination of RPN losses and detection head losses. 

\paragraph{Scene categorization} The benefit of focused attention comes from a pretrained detection model and the scene task itself does not optimize the relation loss. 

\paragraph{Document Classification} For language tasks, we applied
\begin{equation}
    \mathcal{L} = \mathcal{L}_{cls} + \lambda \mathcal{L}_{rel}, \lambda = 0.1
\end{equation}
where $\mathcal{L}_{cls}$ is a standard cross entropy loss for the classification task. 

\section{Implementation Details}
We ran multiple trials (3-5) of our experiments and reported error bars in our results, and observed that the numbers are relatively stable and are reproducible. Furthermore, we plan to release our code upon acceptance of this article. Given that all our datasets are publicly available this will allow other researchers to both reproduce our experiments and use our Focused Attention Network module for their own research.

\subsection{Vision Tasks} We implemented the center-mass cross entropy loss as well as the Focused Attention Module using MxNet. For the Faster R-CNN backbone, we adopted the source code from Relation Networks \cite{hu2017relation}.   

\paragraph{Scene Categorization.} In order to maintain the learned relationship weights from the pre-trained module, which helps encode object co-occurrence context in the aggregation result, we fix the parameters in the convolution backbone, RPN layer and Focused Attention Network module, but make all other layers trainable. Fixed layers are shaded in grey in Figure \ref{fig:net_architecture} (bottom middle).

\subsection{Language Tasks} We implemented the Hierarchical Attention Networks according to \cite{hatt} in Keras with a TensorFlow backend. The word-to-word Focused Attention Network module as well as the center-mass cross entropy loss, are also implemented in the same Keras based framework. 

\subsection{Runtime and machine configuration} 
All our experiments are carried out on a linux machine with 2 Titan XP GPUs, an Intel Core i9 CPU and 64GBs of RAM. The Figures and Tables referred to in the following text are those in the main article. 
\begin{itemize}
    \item \textbf{Figure 4, Relationship Proposal.} For a typical run of Visual Genome Focused Attention Network training, it takes 55 hours for 8 epochs using the above machine configuration.
    \item \textbf{Table 1 Object, Detection.} For a typical run of VOC07 Focused Attention Network training, it takes 4 hours when training for 8 epochs. For a typical run on minicoco , it takes 26 hours using the same setup.
    \item \textbf{Table 2, Scene Categorization.} For a typical run of the MIT67 dataset, it takes 2 hours when training for 8 epochs. 
    \item \textbf{Table 3, Document Classification.} For a typical run of the 20 Newsgroup dataset, it takes 30 minutes for 15 epochs.
\end{itemize}
We also determined that when compared with unsupervised cases of the above experiments, the use of the Focused Attention Network module does not add any noticeable run time overhead.

\section{Additional Visualizations}
\paragraph{Word Importance in a Sentence} In Figure \ref{fig:add_nlp}, 
we provide additional visualizations of the word importance factor in a sentence (defined in Section 4.5 of the main article), using the same format as that used in Figure 2 in the main article.

\begin{table}[h]
    \centering
    \begin{tabular}{c  c}
    \centering
      \includegraphics[width=0.49\textwidth, trim={0cm 0cm 0.3cm 0cm},clip]{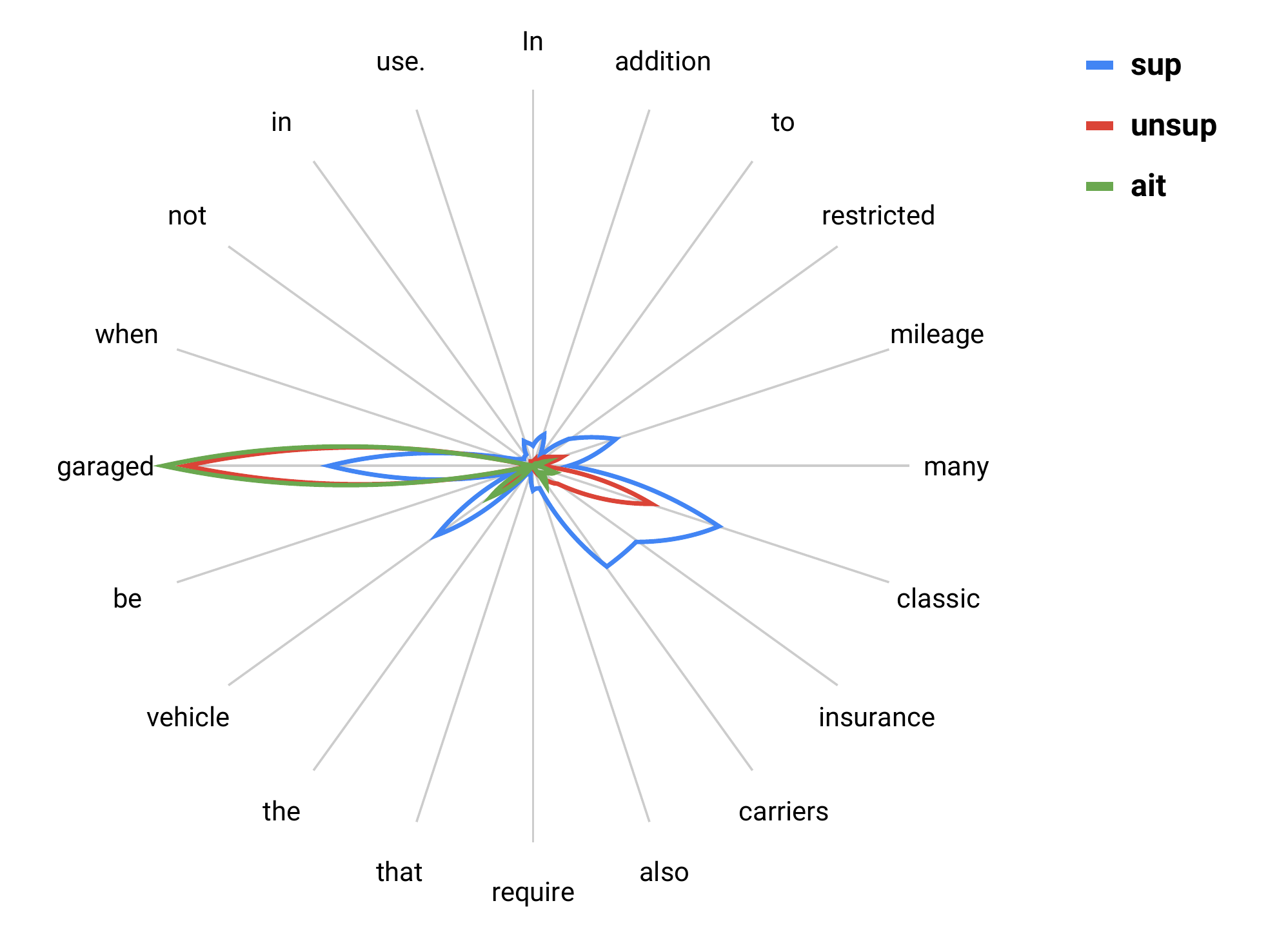} & 
      \includegraphics[width=0.5\textwidth, trim={0cm 0cm 0cm 0cm},clip]{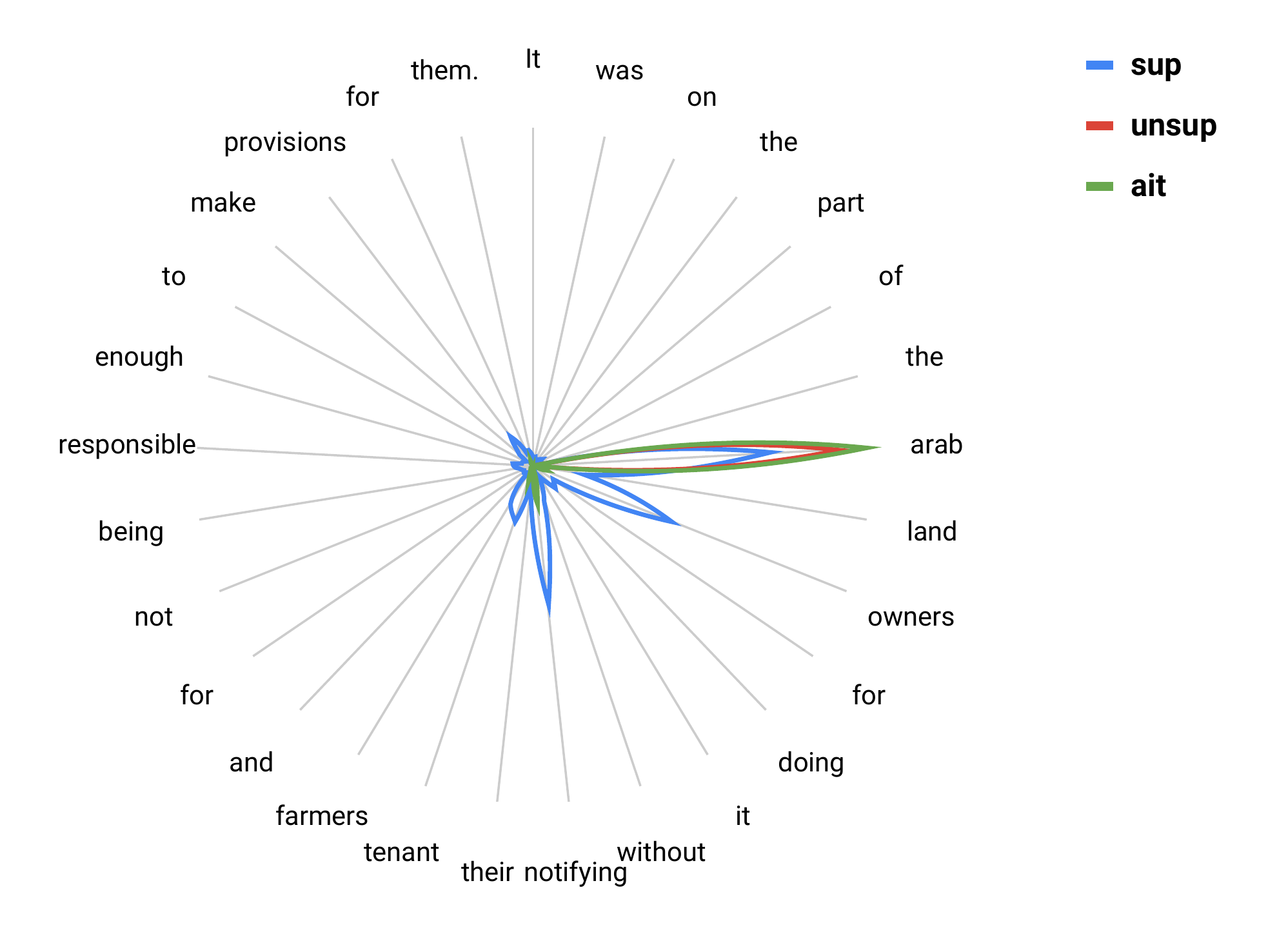} 
    \end{tabular}
    \captionof{figure}{Additional visualization of the word importance factor in a sentence. See Section 3.2 and the caption of Figure 2 of the main article for an explanation.
    }
    \label{fig:add_nlp}
\end{table}

\paragraph{Visual Relationships} We now provide additional qualitative visualizations showing typical relationship weights learned by our method. In Figure \ref{fig:add_vis_rel}, we visualize the predicted relationship on images from the MIT67 dataset, using a pre-trained Focused Attention Network on the minicoco dataset, referred to as FAN-minicoco, as discussed in Section 5.1 of the main article. We compare this with the corresponding unsupervised case, which is similar to Relation Networks \cite{hu2017relation}. 

\begin{table}[h]
    \centering
    \begin{tabular}{c | c}
    Relation Networks \cite{hu2017relation} & Focused Attention Networks \\
    \hline
    \centering
      \includegraphics[width=0.4\textwidth, trim={0 0cm 0 0cm}, clip]{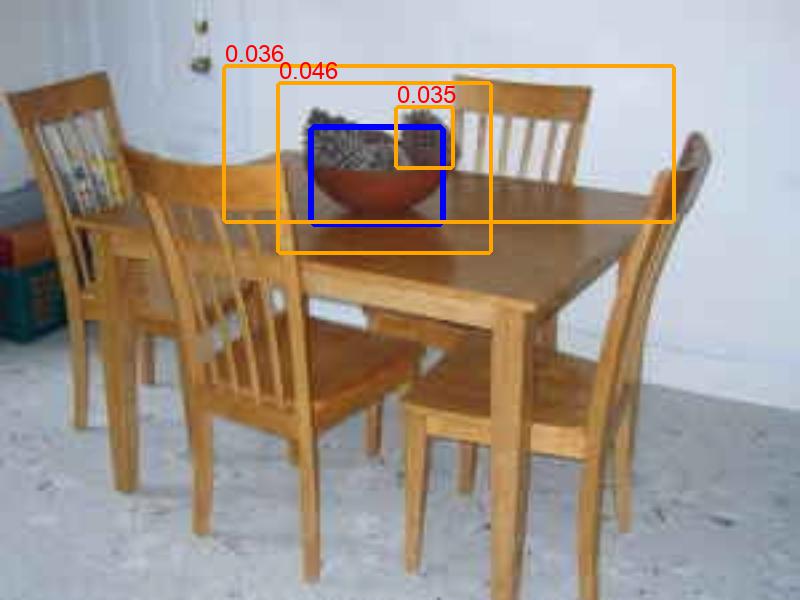} & 
      \includegraphics[width=0.4\textwidth, trim={0 0cm 0 0cm}, clip]{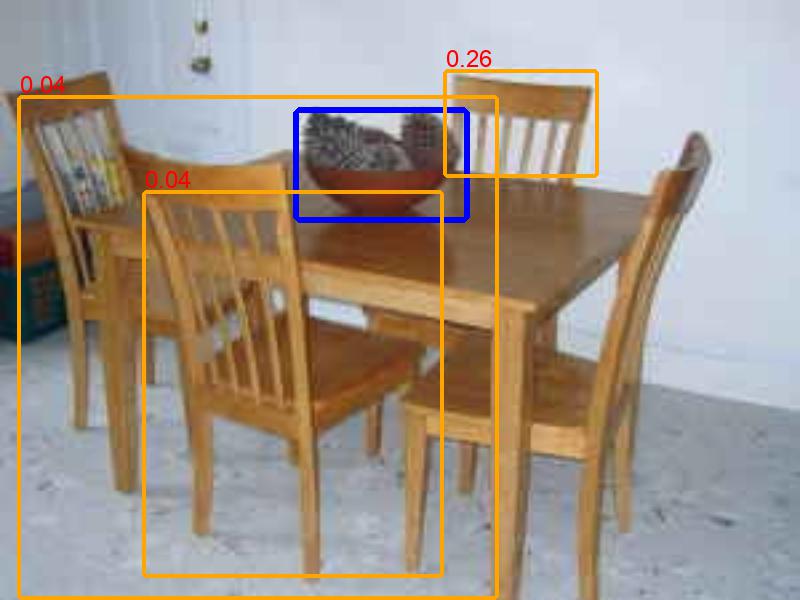} \\
      \includegraphics[width=0.4\textwidth, trim={0 0cm 0 0cm}, clip]{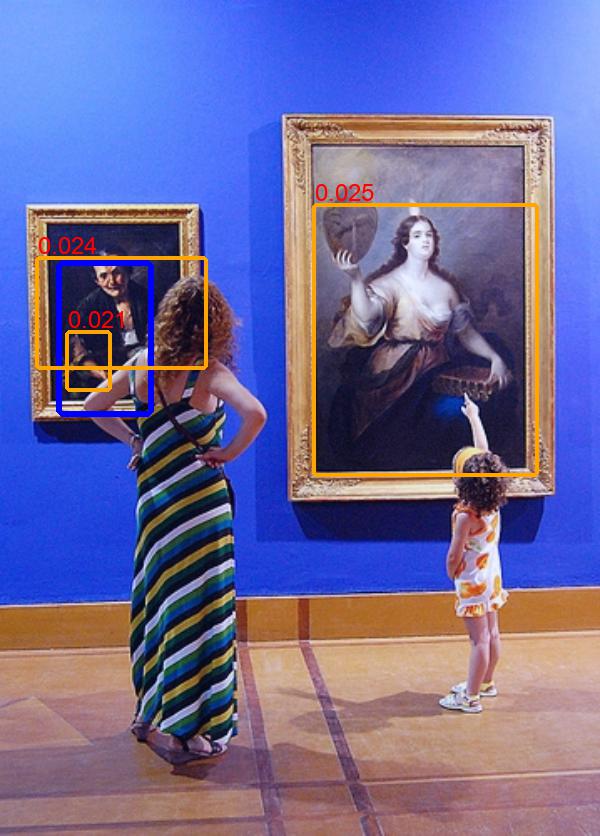} & 
      \includegraphics[width=0.4\textwidth, trim={0 0cm 0 0cm}, clip]{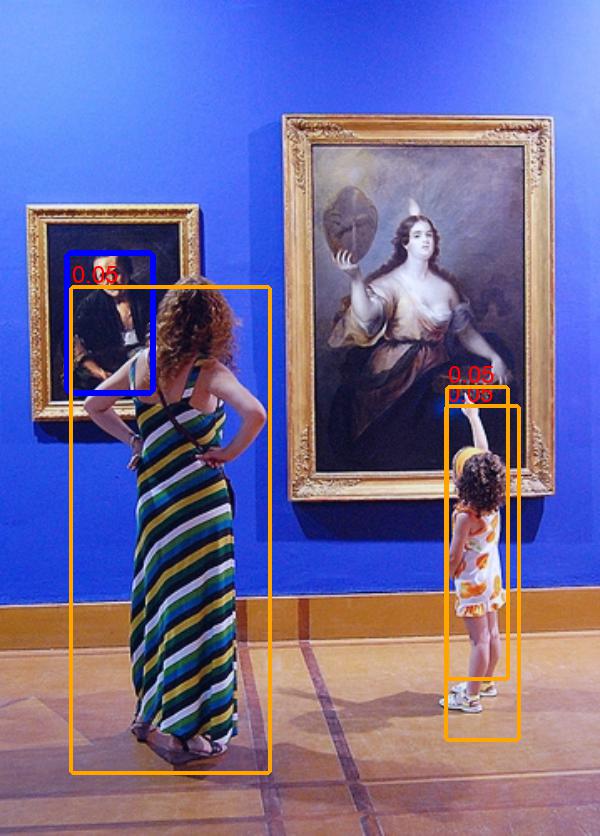} \\
      \includegraphics[width=0.4\textwidth, trim={0 0cm 0 0cm}, clip]{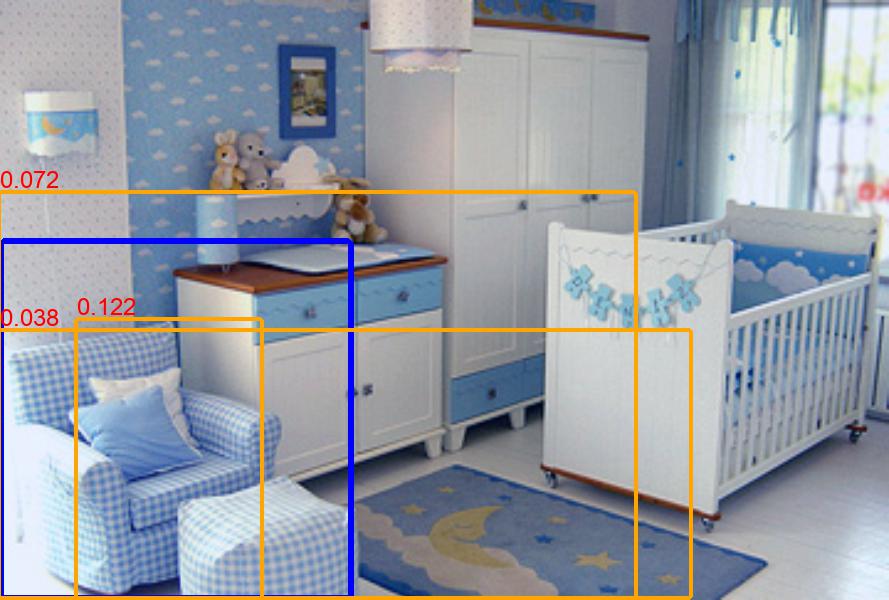} & 
      \includegraphics[width=0.4\textwidth, trim={0 0cm 0 0cm}, clip]{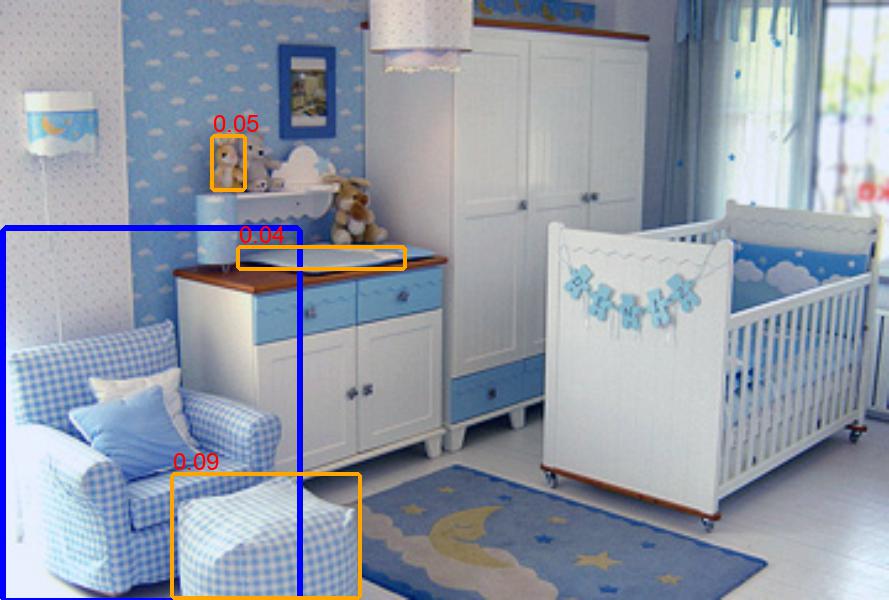} \\
      \includegraphics[width=0.4\textwidth, trim={0 0cm 0 0cm}, clip]{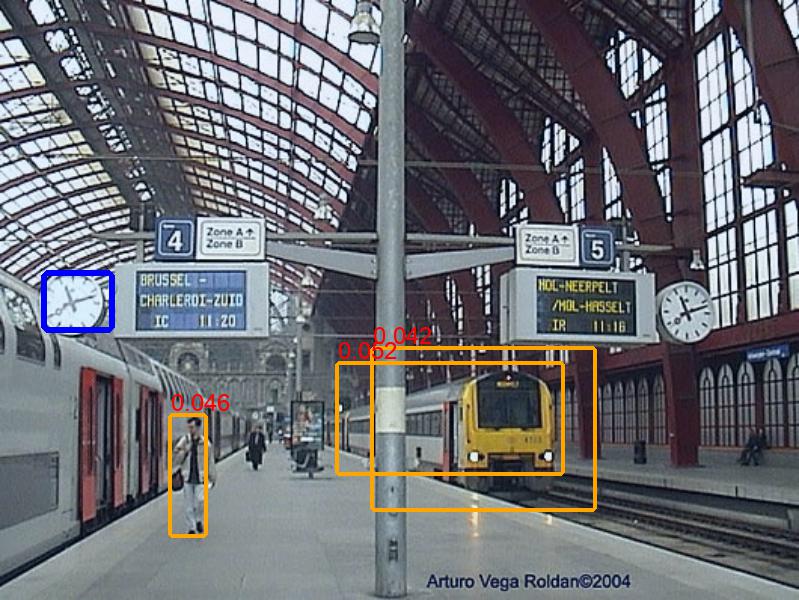} & 
      \includegraphics[width=0.4\textwidth, trim={0 0cm 0 0cm}, clip]{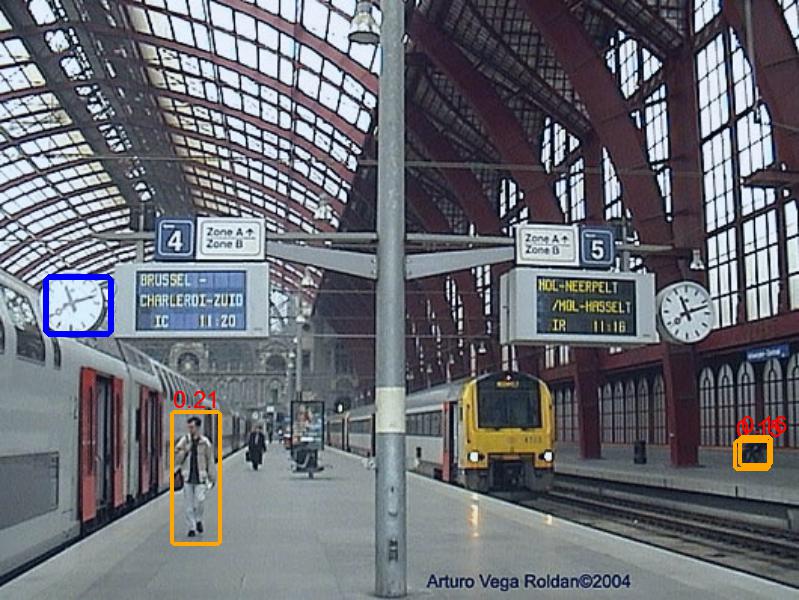} \\

    \end{tabular}
    \captionof{figure}{The visualization of relationships recovered on additional images of the MIT67 dataset. See the caption of Figure 1 of the main article for an explanation.
    }
    \label{fig:add_vis_rel}
\end{table}


\clearpage